# Dimension-Level Intent Fidelity Evaluation for Large Language Models: Evidence from Structured Prompt Ablation

**Gang Peng**

Huizhou Lateni AI Technology Co., Ltd., Huizhou, China

Huizhou University, Huizhou, China

ORCID: 0009-0007-4774-1681

Correspondence: peng@hzu.edu.cn

## ABSTRACT

Holistic evaluation scores capture overall output quality but do not distinguish whether a model reproduced the structural form of a user's request from whether it preserved the user's specific intent. We propose a **dimension-level intent fidelity evaluation framework** — applied here through a structured prompt ablation study across 2,880 outputs spanning three languages, three task domains, and six LLMs — that separately measures structural recovery and intent fidelity for each semantic dimension. This framework reveals a systematic **structural-fidelity split**: among Chinese-language outputs with complete paired scores, 25.7% received perfect holistic alignment scores (GA=5) while exhibiting measurable dimensional intent deficits; among English-language outputs, this proportion rose to 58.6%. Human evaluation confirmed that these split-zone outputs represent genuine quality deficits and that dimensional fidelity scores track human judgements more reliably than holistic scores do. A public–private decomposition of 2,520 ablation cells characterises when models successfully compensate for missing intent and when they fail, while proxy annotation distinguishes prior inferability from default recoverability. A weight-perturbation experiment shows that moderate misalignment is typically absorbed, whereas severe dimensional inversion is consistently harmful. These findings demonstrate that dimension-level intent fidelity evaluation is a necessary complement to holistic assessment when evaluating LLM outputs for user-specific tasks.

Holistic evaluation benchmarks have become central to assessing large language model (LLM) outputs, yet they aggregate all quality dimensions into a single score that may obscure more granular failures. A model that returns a structurally complete response — one that addresses every expected subsection of a task — can nonetheless miss the specific content the user intended, producing output that *looks* correct while being substantively misaligned with the user's request. We term this the **structural-fidelity split** and propose a dimension-level evaluation framework to measure it systematically.

This failure mode is distinct from factual hallucination — the well-studied phenomenon in which models generate information not grounded in their training data or retrieved context[1–3]. Proposed remedies for hallucination — retrieval-augmented generation, reinforcement learning from human feedback, chain-of-thought prompting, and uncertainty quantification — all target the model's knowledge representation[4–7]. A complementary failure mode originates on the *user* side: intent that is present in the user's mind but absent from the prompt. When a user asks for "a competitor analysis for our B2B enterprise clients" and the model returns a structurally valid analysis targeting the wrong audience, no retrieval system can supply the missing intent — because it was never encoded. Factual hallucination and intent-level divergence therefore differ in origin, detection, and mitigation.

This failure mode is also distinct from previously documented LLM-as-judge biases — including position effects, verbosity preferences, and self-familiarity biases[13–15]. Those biases concern *how evaluators compare outputs*; the structural-fidelity split concerns a more fundamental granularity problem, in which the evaluation target itself conflates structural slot-filling and user-specific content fidelity into a single aggregate score.

We address this gap by proposing a **dimension-level intent fidelity evaluation framework** that decouples two layers of alignment previously conflated: **structural recovery** (whether a dimension's slot is filled in the output) and **intent fidelity** (whether the filled content matches the user's actual specification). Using the 5W3H structured prompting protocol[8–10] as a testbed — which decomposes intent into eight dimensions (What, Why, Who, When, Where, How-to-do, How-much, How-feel) — we systematically remove individual dimensions and measure both layers across six models, three languages, and three domains.

This paper makes three contributions. First, we propose the dimension-level intent fidelity evaluation framework and demonstrate through structured prompt ablation that holistic evaluation metrics miss dimensional intent deficits in 25.7–58.6% of outputs with complete paired scores — a discrepancy confirmed by independent human evaluation, in which the same split-zone outputs received mean human scores of 3.12 against LLM scores of 5.0. Second, the framework enables systematic characterisation of compensation asymmetry: recovery varies sharply across domains and dimensions and depends on whether missing content aligns with model training priors; external proxy annotation further reveals that prior inferability and default recoverability are distinct mechanisms that diverge most strongly in high-frequency domains such as travel. Third, the framework identifies a plateau-like weight-tolerance regime in which moderate misalignment is typically absorbed, whereas severe dimensional inversion is consistently harmful — a finding that clarifies what kind of prompt structure matters most for reliable intent fidelity.

---

# RESULTS

## 2.1 Dimensional evaluation reveals intent deficits invisible to holistic metrics

To establish the existence and prevalence of the structural-fidelity split, we conducted a systematic ablation study across 30 tasks and three languages, using six models for Chinese outputs and three models for English and Japanese outputs (Fig. 1a). For each task, one FULL condition (all intent dimensions present) and 7 ablation conditions (each removing exactly one dimension) were evaluated, yielding **2,880 output records** (ZH: 6 models × 30 tasks × 8 conditions = 1,440; EN: 3 models × 30 tasks × 8 = 720; JA: 3 models × 30 tasks × 8 = 720). Each output received both a holistic alignment score (GA, 1–5) and a dimensional intent coverage score (f-ICMw, [0,1]).

The central finding is a systematic divergence between the two metrics. Split-zone prevalence was computed on records with complete paired GA and f-ICMw scores after quality-control filtering; this yielded 1,440 Chinese-language records and 634 English-language records. Among these 1,440 Chinese-language records, **25.7% (370 records) received GA=5 while exhibiting f-ICMw < 0.8** — holistic evaluation awarded full marks to outputs with detectable dimensional deficits (Fig. 1b). In English outputs, this split was even more pronounced: **58.6% of 634 records** fell into this zone. The holistic ceiling is severe: 84.7% of Chinese-language records received the maximum score GA=5 (Fig. 1c).

Three illustrative cases from the split zone confirm that the deficits are semantically meaningful: BZ01/DeepSeek (−how_to_do): GA=5, f-ICMw=0.600 — no SWOT+Porter methodology present despite being specified; BZ02/Gemini (−why): GA=5, f-ICMw=0.600 — strategic rationale absent; TC01 (−who): GA=5, who score=0 — target audience entirely missing.

Human raters assigned substantially lower scores to split-zone outputs. For the 25 split-zone samples — all of which received LLM GA=5 — the human mean GA was **3.120** (95%CI: ±0.306), a gap of **−1.880** versus the LLM judge. Dimensional fidelity (f-ICMw) showed strong human–LLM alignment: Spearman $\rho$ = **0.695** ($p<0.001$), compared with $\rho = 0.251$ ($p=0.053$, n.s.) for holistic GA (Table 1; Fig. 1d). Inter-rater agreement was substantially higher for f-ICMw ($\rho=0.478$, $p<0.001$) than for GA (Cohen's $\kappa=0.006$), consistent with dimensional scoring providing a more stable evaluation signal.

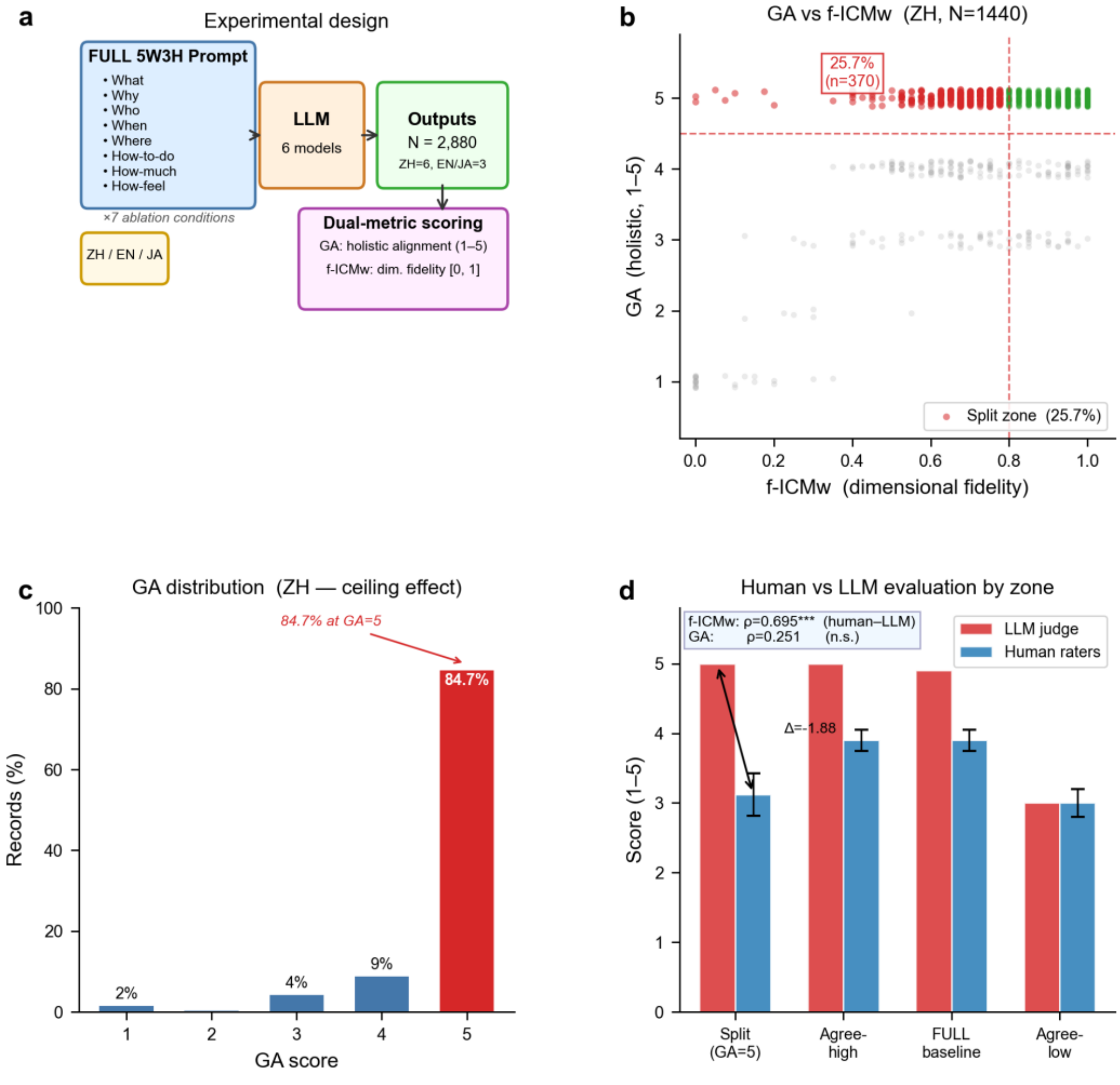

*Figure 1. Dimension-level evaluation and human validation of the structural-fidelity split. (a) Experimental design. (b) GA vs. f-ICMw scatter (ZH, N=1,440); split zone (25.7%) highlighted. (c) GA ceiling effect: 84.7% at GA=5. (d) Human validation: split-zone human mean GA=3.12 vs. LLM GA=5.0.*

## 2.2 Structural recovery systematically diverges from intent fidelity

The s-ICMw ablation hypothesis (H2s: ablating dimension k reduces structural coverage) was supported at high rates across languages — ZH: 83%, EN: 89%, JA: 78%. The f-ICMw hypothesis (H2f: ablation reduces fidelity) was supported at drastically lower rates — ZH: 17%, EN: 0%, JA: 22%. Across 20 matched analysis cells, 19/20 satisfied s-ICMw > f-ICMw support rate (mean $\Delta$ = +0.589), confirming that structural recovery is widespread while fidelity recovery is exceptional. This divergence defines **shallow compensation**: models restore the structural skeleton of missing dimensions using training-prior content, filling slots without matching the user's specific intent.

The structural-fidelity split was observed across all six models spanning the full capability spectrum (frontier: Claude Sonnet 4, GPT-4o; strong: DeepSeek-V3, Qwen-Max; mid-tier: Gemini 2.5 Pro, Kimi). Higher-capability models showed somewhat reduced split rates but did not eliminate the phenomenon, consistent with structural recovery being a fundamental feature of training-prior dynamics rather than a capacity limitation.

### 2.3 Compensation asymmetry and the distinction between inferability and recoverability

We introduce a **public–private decomposition** to characterise when models compensate successfully. A dimension is classified as *public* for a given task-model cell if f-ICMw $\geq 0.7$ when that dimension is ablated; otherwise *private*. Threshold sensitivity was verified at 0.6 and 0.8.

Across 2,520 ablation cells, 795 (31.5%) were public and 1,725 (68.5%) were private (Fig. 2a). Public cells showed a mean f-ICMw **improvement** of 0.118 upon ablation — the **super-recovery effect** (observed in 54.1% of public cells). Private dimensions showed a mean degradation of 0.061. Travel tasks showed the highest public rate (0.514); business (0.227) and technical (0.205) were predominantly private. At the dimension level, How-to-do was most public (0.550) and Who most private (0.156) (Fig. 2b–c).

To examine whether recovery-based publicness reflects genuine prior inferability, we conducted a proxy annotation study (Methods §4.5). GPT-4o and Claude Sonnet 4 independently labeled all 210 task × dimension units blind to model outputs, yielding 0 hard conflicts and Cohen's $\kappa$=0.33. While structural extremes were concordant, overall rank concordance was weak (Spearman $\rho = 0.157$, $p = 0.496$). Travel|where exemplifies the dissociation: recovery public rate = 85% but proxy inferability = 10%, showing that default-recoverability and prior inferability are separable mechanisms (Fig. 2d).

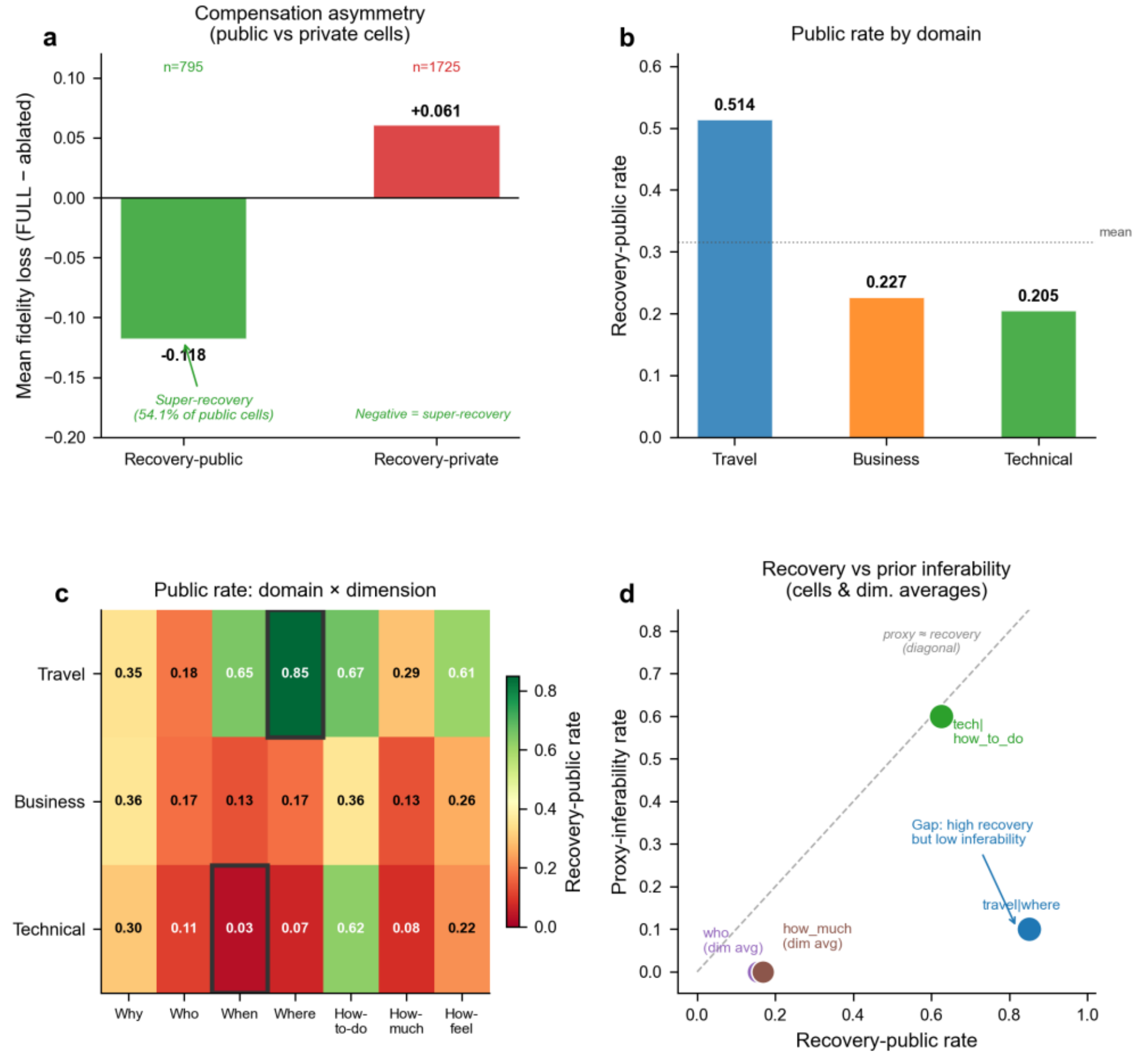

*Figure 2. Compensation asymmetry and the public–private landscape. (a) Mean f-ICMw change: public vs. private cells. (b) Public rate by domain. (c) Domain × dimension heatmap. (d) Recovery-public rate versus proxy-inferability rate for representative cells.*

## 2.4 Severe weight mismatch is robustly harmful under a plateau-like regime

We tested four weight conditions: matched (domain-theoretic prior), uniform (1/8), perturbed (±30% random distortion), and mismatched (high/low-weight dimensions swapped), across 20 tasks, 3 models, 2 domains in two rounds — v2 (240 records) and v3_clean (120 records, leakage-audited).

Both rounds revealed a bimodal structure (Fig. 3; Table 2). Matched, uniform, and perturbed conditions clustered in a **plateau-like robustness zone** (f-ICMw range: 0.959–0.994; within-plateau $\Delta < 0.02$). Mismatched fell to a substantially lower zone (f-ICMw: 0.750–0.861). The Perturbed–Mismatched gap ($\Delta = 0.098$–$0.230$) was approximately **15–25× larger** than the Uniform–Perturbed gap (0.004–0.015). WAS dropped 1.78–1.86 points under mismatched with 100% consistency — the only result with zero exceptions across all cells. The practical implication: *completeness* over *precision* — uniform prompts matched precisely calibrated ones, whereas dimensional inversion was robustly harmful.

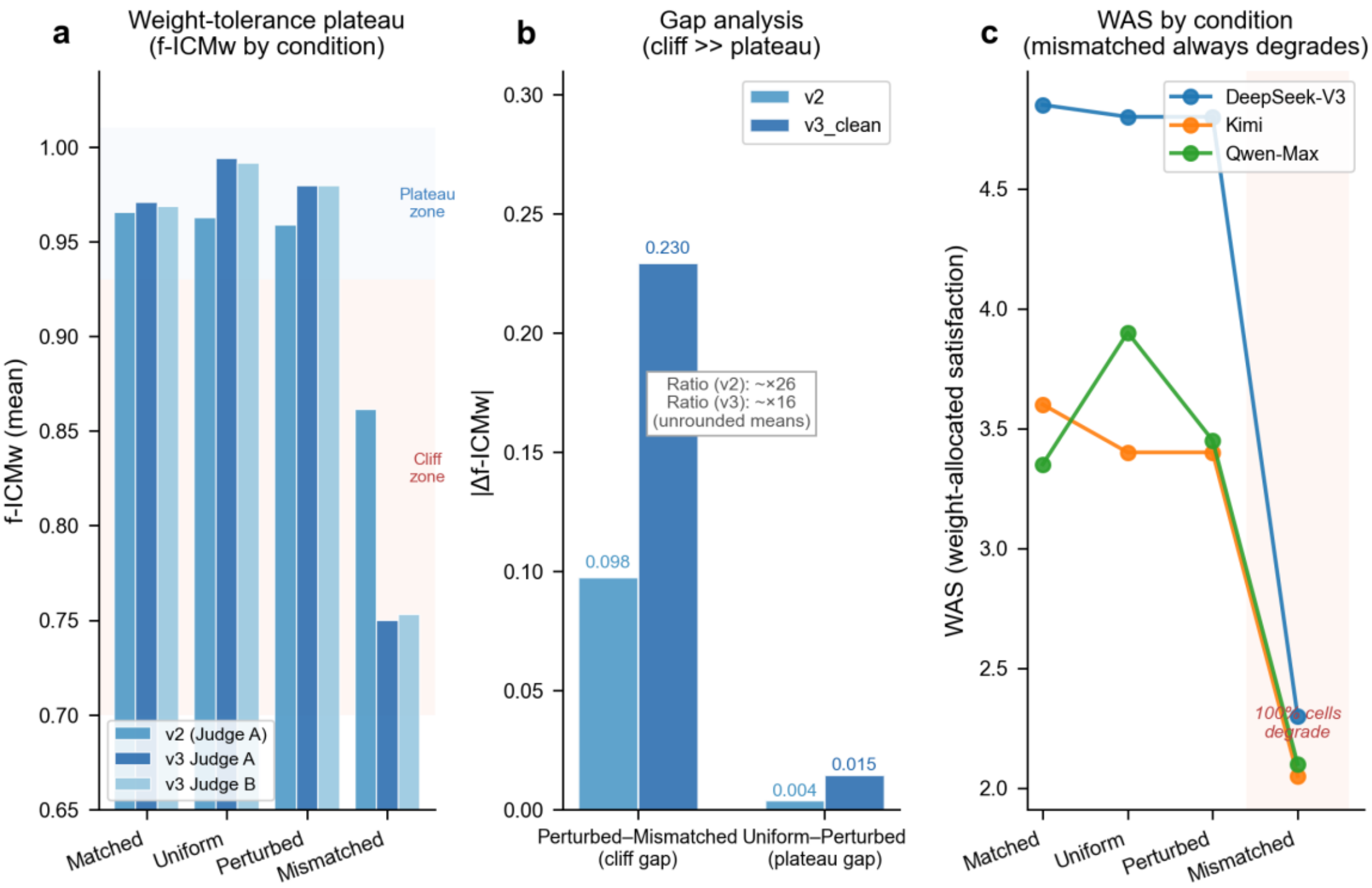

*Figure 3. Weight-tolerance plateau. (a) f-ICMw by condition (v2 and v3_clean). (b) Gap ratio: Perturbed–Mismatched vs. Uniform–Perturbed (~15–25×). (c) WAS drop under mismatched: 100% consistent.*

## 2.5 Cross-language and cross-model robustness

The domain-level ordering of public rates (travel > business ≈ technical) and dimension-level ordering (How-to-do most public, Who most private) were preserved across ZH, EN, and JA. EN outputs showed higher overall public rates than ZH, consistent with English-dominant training data providing stronger English-language priors. JA showed intermediate rates with model-specific variation (Supplementary Figure S1).

Model-level public rates were heterogeneous rather than monotonically ordered by nominal capability (overall Spearman $\rho = -0.029$), suggesting that recovery is shaped not only by model strength but also by language-specific priors and domain coverage. Within English-language data, a partial ordering was observed (Claude Sonnet 4: 0.271; GPT-4o: 0.238; Gemini 2.5 Pro: 0.224) but should be treated as exploratory given the limited model set per language.

---

## DISCUSSION

The structural-fidelity split defines an alignment failure mode distinct from factual hallucination[11]. In shallow compensation, the model generates *plausible, factually grounded content* misaligned with the *specific user intent* — because the user's intent was never encoded in the prompt. Model-side interventions cannot reliably recover intent that never entered the channel. The structural layer (was a "who" section present?) is consistently recovered; the fidelity layer (was it the *right* "who"?) is not. Human evaluation confirms this directly — LLM judges and human raters converge on dimensional fidelity (ρ=0.695) but diverge on holistic assessment (ρ=0.251, n.s.).

The complementary role of dimensional evaluation is to provide measurement resolution that holistic metrics do not. A single alignment score aggregates all quality dimensions into one number, losing the dimensional resolution required to detect partial fidelity. The ceiling effect (84.7% at GA=5) is a symptom of this aggregation: holistic metrics collapse quality variation that dimensional analysis preserves. This phenomenon extends beyond our specific judge: MT-Bench, Chatbot Arena, HELM, and related preference-based benchmarks[13,16,17,18] rely on holistic comparisons that may be insensitive to this failure mode unless supplemented with dimension-level checks.

The empirical findings can be interpreted under a lightweight interpretive framework, **Intent Signal Theory (IST)^8–10^**. Using a communication-channel analogy inspired by Shannon[12], IST treats human–AI interaction as a process in which user intent is encoded into a prompt and partially recovered in the output. *Public* intent components are compatible with model priors and can often be reconstructed; *private* components depend on user-specific constraints and are substantially harder to recover when absent. The framework explains why super-recovery occurs for public dimensions and why severe weight inversion is more damaging than moderate weight noise.

The proxy annotation study further illuminates IST by distinguishing *prior inferability* (gold content predictable from task conventions) from *default-recoverability* (any plausible default achieves partial-match scores). Travel dimensions exemplify the dissociation: high default-recoverability (travel|where = 85%) but low prior inferability (proxy = 10%). *Who* and *how_much* — 0% proxy-inferable — are precisely the dimensions where f-ICMw is most diagnostic. We present IST as a conceptual framework; formal information-theoretic development remains future work.

Three actionable recommendations follow. First, *structured encoding over blank prompting*: uniform weighting outperforms dimensional omission far more than it underperforms precisely calibrated weighting. Second, *dimensional metrics in evaluation*: instruction-following evaluation for user-specific tasks should include dimensional coverage metrics alongside holistic scores. Third, *private-dimension prioritisation*: for high-complexity tasks, explicit specification of Who and How-much yields disproportionate returns because these cannot be recovered by model priors.

Several limitations bound the current findings. The 30-task sample limits statistical power for cross-domain rank comparisons. The weight experiment covers three Chinese-language models only. The public–private classification threshold is partially circular; proxy annotation partially addresses this but reveals additional conceptual complexity. Human evaluation used 60 samples with two raters; larger-scale validation remains future work. The experiments are black-box behavioural; mechanistic analysis would provide causal evidence complementing the current correlational findings.

## METHODS

### 4.1 Task design

Thirty task scenarios were constructed across three domains: travel planning (TR01–TR10), business analysis (BZ01–BZ10), and technical documentation (TC01–TC10). Each task was encoded as a complete 5W3H structured prompt (FULL condition) containing all eight intent dimensions. Chinese (ZH) versions were primary; English (EN) and Japanese (JA) translations were prepared by bilingual authors and verified for dimensional equivalence. Task-conditioned weight vectors w(T) were assigned based on domain-specific importance priors, derived from the PPS framework[8–10].

### 4.2 Ablation protocol

For each task, 8 conditions were constructed: one FULL condition and 7 ablation conditions (one per ablatable dimension; What is anchored as task identifier and excluded from ablation). Ablated prompts preserved all non-target dimensions verbatim. Models received no indication that a dimension was removed. Total records: ZH 1,440 (6 models × 30 tasks × 8 conditions), EN 720, JA 720.

**Models:** Claude Sonnet 4, GPT-4o (frontier); DeepSeek-V3, Qwen-Max (strong); Gemini 2.5 Pro, Kimi (mid). API identifiers: claude-sonnet-4-20250514, gpt-4o, deepseek-chat, qwen-max, gemini-2.5-pro, moonshot-v1-32k. The weighting experiment used DeepSeek-V3, Qwen-Max, and Kimi.

Model–language coverage reflects a two-phase design. Phase 1 used three Chinese-developed models on ZH tasks as a confirmatory pilot. Phase 2 extended to three internationally-developed models across ZH, EN, and JA to assess cross-model and cross-linguistic generalisability. All six models contribute to ZH; EN and JA are covered by the international-model set only.

### 4.3 Evaluation metrics

**Goal Alignment (GA):** Holistic 1–5 score, judge blind to ablation condition. **s-ICMw (Structural Intent Coverage):** Per-dimension binary-ternary score {0, 0.5, 1} for whether a dimension's structural slot is present. **f-ICMw (Fidelity Intent Coverage):** Per-dimension score {0, 0.5, 1} against the FULL prompt specification as gold reference; f-ICMw = $\Sigma\ w_i \cdot f_i$. **DS (Deficiency Signature):** {0, 0.5, 1} scoring of whether dimension-k removal caused the expected failure pattern. Primary judge: DeepSeek-V3 (unified across all 2,880 records). Three-pass scoring protocol: Pass 1 (GA) → Pass 2 (ICMw) → Pass 3 (DS), with progressive information disclosure to prevent judge contamination.

### 4.4 Public/private classification

Ablation cell (task × model × removed_dim) classified as public if f-ICMw(dim=k removed) ≥ 0.7; private otherwise. Threshold sensitivity verified at 0.6 and 0.8. Classification applied to all 2,520 ablation cells (ZH: 6 models × 30 tasks × 7 = 1,260; EN+JA: 3 models × 30 tasks × 7 × 2 = 1,260). All cells returned valid f-ICMw scores and were included.

### 4.5 External proxy annotation of prior inferability

GPT-4o and Claude Sonnet 4 independently labeled all 210 task × dimension units (30 tasks × 7 ablatable dimensions) as public, private, or mixed, based solely on task title and domain context — blind to all model outputs and ablation results. Labels merged under a B+ conservative rule: both-public → public; both-private → private; any disagreement → mixed; zero public–private conflicts observed. Inter-model agreement: 55.2% exact, Cohen's κ=0.33.

## 4.6 Weight experiment

**Conditions:** Matched (domain-theoretic w(T)), uniform (1/8 per dimension), perturbed (±30% random distortion from matched), mismatched (highest-weight and lowest-weight dimensions swapped). **Rounds:** v2 (240 records: 20 tasks × 4 conditions × 3 models × 2 domains), v3_clean (120 records: 10-task leakage-audited subset). PRS audit confirmed textual weight fidelity for all 40 prompts (100% pass). WAS (Weight-Allocated Satisfaction) = weighted holistic satisfaction using applied weight vector.

## 4.7 Statistical methods

Spearman rank correlation (ρ) for H2 support assessment with 10,000-permutation tests. Gini coefficient for weight dispersion. Dual-judge agreement reported as same-band rate and mean absolute difference. Effect sizes: mean Δ with 95% CI. All percentages are exact counts.

## 4.8 Human evaluation

Stratified sample: N=60 outputs (25 split-zone: GA=5, f-ICMw<0.8; 15 agree-high: GA=5, f-ICMw≥0.9; 10 FULL-baseline; 10 agree-low: GA≤3). Two independent human raters (Rater A: familiar with 5W3H; Rater B: no prior exposure) scored each output on GA and per-dimension ICM, blind to condition and LLM scores. Split-zone human mean GA: 3.120 (95% CI: ±0.306) vs. LLM GA=5.0 (Δ=−1.880). Human raters were research collaborators who provided informed consent; all materials were anonymised model outputs.

## 4.9 Relationship to prior work

This paper builds on three prior studies on the 5W3H/PPS framework[8–10] reporting condition-level structured-versus-unstructured prompting comparisons. The present manuscript addresses a distinct question — whether standard evaluation metrics can detect dimension-level intent deficits — and introduces the dimension-level evaluation framework, f-ICMw scoring, public–private decomposition, human validation, proxy inferability annotation, and weight-perturbation experiments as original contributions not reported elsewhere. These analyses are not under consideration at any other journal.

---

**TABLE 1 Human Evaluation Results**

| Zone | N | Human-A GA | Human-B GA | Avg GA | LLM GA | Δ(Avg−LLM) | Human ICMw | LLM ICMw |
|---|---|---|---|---|---|---|---|---|
| Split (GA=5, f-ICMw<0.8) | 25 | 3.04 | 3.20 | 3.12 | 5.00 | −1.88 | 0.598 | 0.702 |
| Agree-high (GA=5, f-ICMw≥0.9) | 15 | 3.67 | 4.13 | 3.90 | 5.00 | −1.10 | 0.759 | 0.895 |
| Full-baseline (FULL condition) | 10 | 4.20 | 3.60 | 3.90 | 4.90 | −1.00 | 0.719 | 0.852 |
| Agree-low (GA≤3) | 10 | 2.40 | 3.60 | 3.00 | 3.00 | 0.00 | 0.616 | 0.720 |
| **Summary** (N=60) | | | | | | | **ρ(ICMw)=0.695** | **ρ(GA)=0.251** |

Spearman ρ between human-average and LLM scores across all 60 outputs: f-ICMw $\rho=0.695$ ($p<0.001$); GA $\rho=0.251$ ($p=0.053$, n.s.). Inter-rater f-ICMw: $\rho=0.478$ ($p<0.001$); inter-rater GA: Cohen's $\kappa=0.006$. *** $p < 0.001$.

**TABLE 2 Weight-Tolerance Experiment Summary**

| Condition | Avg f-ICMw (A) | Avg f-ICMw (B) | Avg WAS | f-band agr. | Zone |
|---|---|---|---|---|---|
| Matched | 0.971 | 0.969 | 4.03 | 100% | plateau |
| Uniform | 0.994 | 0.992 | 3.96 | 100% | plateau |
| Perturbed | 0.980 | 0.980 | 3.97 | 100% | plateau |
| **Mismatched** | **0.750** | **0.753** | **2.17** | 100% | **cliff** |

Values are means across 6 cells (2 domains × 3 models). f-band agreement: |Judge A − Judge B| ≤0.10. WAS = weight-allocated satisfaction score. Per-domain per-model breakdown in Supplementary Table S1.

---

## COMPETING INTERESTS

G.P. is the creator of the 5W3H/PPS structured prompting framework and co-founder of Huizhou Lateni AI Technology Co., Ltd., which develops software tools related to structured prompt authoring. This potential competing interest has been disclosed. No other competing interests are declared.

## DATA AVAILABILITY

Data required to reproduce all figures and tables are provided as source data files accompanying this submission. The task prompts, ablation conditions, anonymised model outputs, scoring files, human-evaluation scores, proxy-annotation labels, and analysis scripts are publicly available at https://github.com/PGlarry/prompt-protocol-specification/tree/main/dataset/data/structural_fidelity_split.

## CODE AVAILABILITY

Analysis code for ablation scoring, public–private classification, proxy annotation aggregation, weight-experiment analysis, and figure generation is publicly available at https://github.com/PGlarry/prompt-protocol-specification/tree/main/dataset/data/structural_fidelity_split. The 5W3H/PPS structured prompting platform used to generate experimental prompts is available at https://www.lateni.com.

---

## REFERENCES

1. Huang, L. et al. A survey on hallucination in large language models. *ACM Comput. Surv.* 57, 1–38 (2023).

2. Ji, Z. et al. Survey of hallucination in natural language generation. *ACM Comput. Surv.* 55, 1–38 (2023).

3. Rawte, V. et al. A survey of hallucination in large foundation models. Preprint at *arXiv:2309.05922* (2023).

4. Lewis, P. et al. Retrieval-augmented generation for knowledge-intensive NLP tasks. *NeurIPS* (2020).

5. Ouyang, L. et al. Training language models to follow instructions with human feedback. *NeurIPS* (2022).

6. Wei, J. et al. Chain-of-thought prompting elicits reasoning in large language models. *NeurIPS* (2022).

7. Farquhar, S. et al. Detecting hallucinations in large language models using semantic entropy. *Nature* 630, 625–630 (2024).

8. Peng, G. Evaluating 5W3H Structured Prompting for Intent Alignment in Human–AI Interaction. Preprint at *arXiv:2603.18976* (2026).

9. Peng, G. Does Structured Intent Representation Generalize? A Cross-Language, Cross-Model Empirical Study of 5W3H Prompting. Preprint at *arXiv:2603.25379* (2026).

10. Peng, G. Structured Intent as a Protocol-Like Communication Layer. Preprint at *arXiv:2603.29953* (2026).

11. Xu, Z., Jain, S. & Kankanhalli, M. Hallucination is inevitable. Preprint at *arXiv:2401.11817* (2024).

12. Shannon, C. E. A mathematical theory of communication. *Bell Syst. Tech. J.* 27, 379–423 (1948).

13. Zheng, L. et al. Judging LLM-as-a-judge with MT-bench and Chatbot Arena. *NeurIPS* 36 (2023).

14. Dubois, Y. et al. Length-controlled AlpacaEval. Preprint at *arXiv:2404.04475* (2024).

15. Panickssery, A. et al. LLM evaluators recognize and favor their own generations. Preprint at *arXiv:2404.13076* (2024).

16. Zhou, J. et al. Instruction-following evaluation for large language models. Preprint at *arXiv:2311.07911* (2023).

17. Liang, P. et al. Holistic evaluation of language models. *Trans. Mach. Learn. Res.* (2023).

18. Chang, Y. et al. A survey on evaluation of large language models. *ACM Trans. Intell. Syst. Technol.* 15, 1–45 (2024).

# SUPPLEMENTARY INFORMATION

## Supplementary Note 1 — Ablation Cell Quality-Control Summary

Of 2,520 ablation cells (3 domains × 840 cells per domain), all 2,520 passed quality control with valid per-dimension f-ICMw scoring. No cells were excluded.

| Domain | Possible cells | Valid cells | Excluded | Exclusion rate |
|---|---|---|---|---|
| Business | 840 | 840 | 0 | 0.0% |
| Technical | 840 | 840 | 0 | 0.0% |
| Travel | 840 | 840 | 0 | 0.0% |
| **Total** | **2,520** | **2,520** | **0** | **0.0%** |

## Supplementary Table S1 — Full Weight-Tolerance Results (v3_clean)

| Domain | Model | Condition | f-ICMw (A) | f-ICMw (B) | WAS | f-band agr. | Zone |
|---|---|---|---|---|---|---|---|
| Business | DeepSeek-V3 | Matched | 0.996 | 0.996 | 5.00 | same | plateau |
| Business | DeepSeek-V3 | Uniform | 0.996 | 0.996 | 5.00 | same | plateau |
| Business | DeepSeek-V3 | Perturbed | 0.958 | 0.958 | 4.83 | same | plateau |
| Business | DeepSeek-V3 | Mismatched | 0.817 | 0.817 | 2.67 | same | cliff |
| Business | Qwen-Max | Matched | 0.996 | 0.996 | 4.00 | same | plateau |
| Business | Qwen-Max | Uniform | 0.996 | 0.983 | 4.00 | same | plateau |
| Business | Qwen-Max | Perturbed | 0.988 | 0.988 | 3.33 | same | plateau |
| Business | Qwen-Max | Mismatched | 0.946 | 0.933 | 2.17 | same | cliff† |
| Business | Kimi | Matched | 0.996 | 0.996 | 3.67 | same | plateau |
| Business | Kimi | Uniform | 0.979 | 0.983 | 4.00 | same | plateau |
| Business | Kimi | Perturbed | 0.988 | 0.988 | 3.67 | same | plateau |
| Business | Kimi | Mismatched | 0.988 | 1.000 | 2.17 | same | cliff† |
| Technical | DeepSeek-V3 | Matched | 0.988 | 0.988 | 5.00 | same | plateau |
| Technical | DeepSeek-V3 | Uniform | 1.000 | 1.000 | 4.75 | same | plateau |
| Technical | DeepSeek-V3 | Perturbed | 1.000 | 1.000 | 5.00 | same | plateau |
| Technical | DeepSeek-V3 | Mismatched | 0.375 | 0.469 | 2.00 | same | cliff |
| Technical | Qwen-Max | Matched | 0.856 | 0.844 | 3.00 | same | plateau |
| Technical | Qwen-Max | Uniform | 0.994 | 0.988 | 3.50 | same | plateau |
| Technical | Qwen-Max | Perturbed | 0.975 | 0.975 | 3.50 | same | plateau |
| Technical | Qwen-Max | Mismatched | 0.569 | 0.500 | 2.00 | same | cliff |
| Technical | Kimi | Matched | 0.994 | 0.994 | 3.50 | same | plateau |
| Technical | Kimi | Uniform | 1.000 | 1.000 | 2.50 | same | plateau |
| Technical | Kimi | Perturbed | 0.969 | 0.969 | 3.50 | same | plateau |
| Technical | Kimi | Mismatched | 0.806 | 0.800 | 2.00 | same | cliff |

f-band agreement: $|\text{Judge A} - \text{Judge B}| \leq 0.10$ (100% across all 24 cells). WAS = weight-allocated satisfaction score. †Business mismatched (Qwen-Max/Kimi): high f-ICMw because structural recovery compensates; WAS drop confirms functional degradation.

## Supplementary Figure S1 — Cross-Language Structural-Fidelity Support Rates

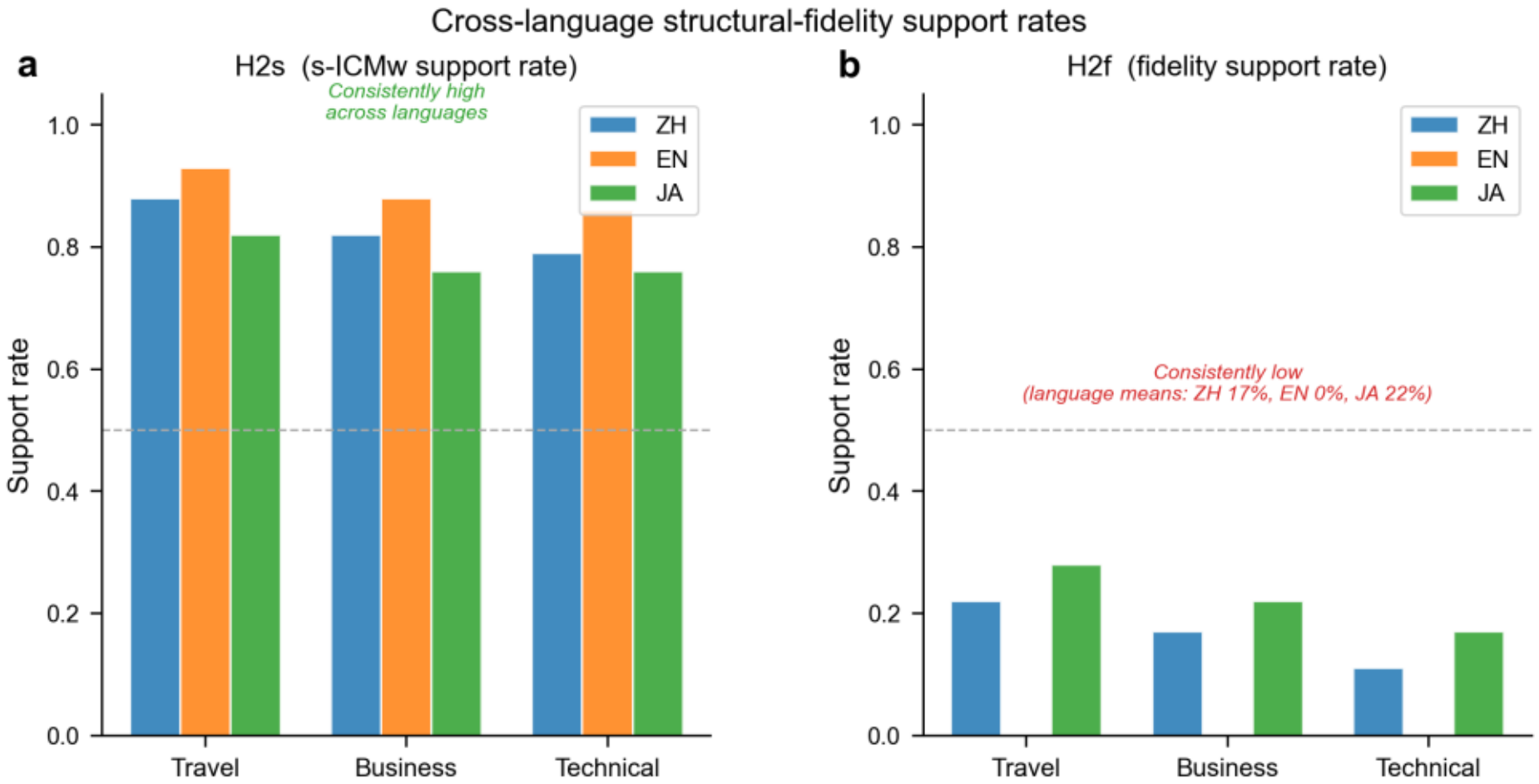

*Supplementary Figure S1. H2s and H2f support rates by language × domain. s-ICMw (structural) consistently high; f-ICMw (fidelity) consistently low. Demonstrates cross-linguistic generalisation.*

## Supplementary Box 1 — Representative FULL-condition Prompt (Travel Domain)

The following is a verbatim example of a FULL-condition structured prompt generated by the lateni.com PPS platform (PPS Standard v1.0.0). All six models received prompts in this format for the FULL condition; ablated variants removed individual dimension blocks while retaining all remaining dimensions and the execution trigger unchanged. The sha256 hash covers the complete prompt text for post-hoc verification.

PPS Standard: v1.0.0 | Prompt ID: b29e8c93351d | Created: 2026-05-02T03:48:41

sha256: af810df2f534aa7d7d935c2c27fb9e0ba4a76e7e9e7818b19ddcb29e8c93351d

Objective (What): Five-Day Travel Guide to New York

Reason (Why): To provide a comprehensive and efficient itinerary for experiencing New York City's iconic landmarks, diverse culture, and vibrant neighborhoods within a limited five-day timeframe; to help travelers maximize their trip by balancing must-see attractions with hidden gems and local experiences.

Role (Who): First-time and returning tourists seeking a structured yet flexible itinerary; solo travelers, couples, families with older children, and small groups of friends; travel enthusiasts interested in culture, history, food, and urban exploration.

Schedule (When): Best suited for spring (April to June) and fall (September to November) when weather is mild and crowds are moderate; guide accounts for seasonal variations, holiday peaks, and weekday vs. weekend crowd patterns.

Location (Where): Focusing on Manhattan as the primary hub, with excursions to Brooklyn, Queens, and other boroughs; covers major zones including Midtown, Lower Manhattan, Upper East/West Sides, SoHo, Greenwich Village, and DUMBO.

Method (How to do): Organise the guide into five daily themes; include a mix of scheduled activities and free time; provide practical tips for booking tickets in advance, using city passes, and managing time between attractions.

Metrics (How much): A five-day itinerary covering 15–20 major attractions, 10–15 restaurants, 5–8 neighbourhoods; estimated budget $1,500–$3,000 per person (excluding flights and accommodation); guide length ~3,000–4,000 words.

Outcome (How feel): Exciting and immersive; practical and reassuring; inspiring and aspirational; balanced between adventurous exploration and comfortable pacing.

Please execute the task according to the above content.